\documentclass[letterpaper, 10 pt, conference]{ieeeconf}
\IEEEoverridecommandlockouts % for \thanks
\usepackage{graphicx} % for \includegraphics
\usepackage{booktabs}     
\usepackage{cite}
\usepackage{float}
\graphicspath{{./pdf/}} % load figure within subfolder ./pdf/
\DeclareGraphicsExtensions{.pdf}

\usepackage[export]{adjustbox} % for \adjincludegraphics

\usepackage[cmex10]{amsmath} % assumes amsmath package installed

\usepackage{bm} % for \bm

\usepackage{epsfig} % for postscript graphics files

\usepackage{times} % assumes new font selection scheme installed

\usepackage{color} % for \color

\usepackage{amssymb}  % assumes amsmath package installed

\newcommand{\mytilde}{\raise.17ex\hbox{$\scriptstyle\mathtt{‌​\sim}$}}

\usepackage{graphicx,wrapfig,amsfonts,multirow,bbm,algorithm,algpseudocode}
\title{\LARGE \bf
	Automatic Gesture Recognition in Robot-assisted Surgery\\ with Reinforcement Learning and Tree Search 
}

\author{Xiaojie Gao, Yueming Jin, Qi Dou, and Pheng-Ann Heng% <-this % stops a space
	\thanks{This work was partially supported by HK RGC TRS project T42-409/18-R, and a grant from the National Natural Science Foundation of China (Project No. U1813204) and CUHK T Stone Robotics Institute.}
	\thanks{X. Gao, Y. Jin, Q. Dou and P. A. Heng are with the Department of Computer Science and Engineering, The Chinese University of Hong Kong. Q. Dou and P. A. Heng are also with the CUHK T Stone Robotics Institute. \emph{Corresponding author at: qdou@cse.cuhk.edu.hk (Qi Dou).}}
}

\begin{document}

\maketitle
\thispagestyle{empty}
\pagestyle{empty}

\begin{abstract}
Automatic surgical gesture recognition is fundamental for improving intelligence in robot-assisted surgery, such as conducting complicated tasks of surgery surveillance and skill evaluation. However, current methods treat each frame individually and produce the outcomes without effective consideration on future information. In this paper, we propose a framework based on reinforcement learning and tree search for joint surgical gesture segmentation and classification. An agent is trained to segment and classify the surgical video in a human-like manner whose direct decisions are re-considered by tree search appropriately. Our proposed tree search algorithm unites the outputs from two designed neural networks, i.e., policy and value network. With the integration of complementary information from distinct models, our framework is able to achieve the better performance than baseline methods using either of the neural networks. For an overall evaluation, our developed approach consistently outperforms the existing methods on the suturing task of JIGSAWS dataset in terms of accuracy, edit score and F1 score. Our study highlights the utilization of tree search to refine actions in reinforcement learning framework for surgical robotic applications.

\end{abstract}

%======================================================
\section{INTRODUCTION}

Robot-assisted surgery facilitates surgeons to perform a variety of complex operations in minimally invasive surgery and improves the precision of surgical manipulation in the meanwhile. For example, the \textit{da Vinci} surgical system is designed to assist certain surgery and widely used in 
%minimally invasive 
nowadays clinical procedures, with a large amount of video visual and kinematic data recorded \cite{van2019weakly}.
Towards an intelligent operating theatre, developing data-driven methods which can learn to recognize the surgical gestures is a fundamental task. The goal is to segment a sequence of surgical operations given in video format, 
%and/or other sensor information and 
i.e., classifying each frame into a specific type of surgical gesture, such as \textit{positioning needle}, \textit{orienting needle}, and \textit{pulling the suture}, etc. 

Automatically recognizing the robotic gestures in surgical process plays an important role for surgery surveillance \cite{dipietro2019segmenting}, automatic skill assessment \cite{reiley2009task,varadarajan2009data,poursartip2018analysis}, and surgery training \cite{ahmidi2017dataset}. Identifying which action is being operated is also crucial for developing the context-aware theatre \cite{dergachyova2016automatic} and autonomous robotic surgery systems \cite{preda2016cognitive}. These applications help reduce the mental cognitive workload of surgeons and improve reliability and safety of the robot-assisted surgery.
%by automatically identifying and evaluating actions. 
However, the development of automatic surgical gesture recognition method is challenging, as the gesture usually includes complex multi-step actions and sometimes intricate maneuvers \cite{tsai2019transfer}. Also, variability in users' manipulation habits and proficiency makes the problem even more complicated. 

Some studies have been conducted for surgical recognition tasks, ranging from phase recognition to fine-grained gesture and action recognition.
Classical approaches have been based on statistical models and unsupervised learning methods, e.g., Hidden Markov Model (HMM) was exploited for automatic phase recognition \cite{tran2017phase}. A sparse HMM was proposed to improve the expressive power of discrete or Gaussian observations \cite{tao2012sparse}. Fusing video and kinematic data, a combined Markov/semi-Markov conditional random field (MsM-CRF) model was exploited for joint segmentation and recognition of surgical gestures \cite{tao2013surgical}. In \cite{mavroudi2018end}, a temporal CRF model was combined with a frame-level representation based on discriminative sparse coding.
However, these methods produce suboptimal outcomes because of either loss of long-term dependency or severe over-segmentation problems \cite{liu2018deep}. Gaussian Mixture Model (GMM) initialized by the k-means clustering algorithm was used to estimate segmentation points \cite{lee2015autonomous}. A hierarchical Dirichlet Process GMM was proposed to learn the segmentation criteria \cite{krishnan2018transition}. To avoid tedious parameter tuning, a GMM-based algorithm was designed under weak supervision \cite{van2019weakly}. However, the manual feature extraction used in these methods is relatively subjective with limited representation capability.

For better feature extraction, methods using deep learning (DL) techniques have achieved impressive results in surgical recognition. Long Short Term Memory (LSTM) network was used to maintain the temporal information among frames \cite{dipietro2016recognizing}. Designed to better extract low-level features, segmental spatiotemporal convolutional neural network (Seg-ST-CNN) outperformed temporal models like LSTM \cite{lea2016segmental}. In \cite{lea2016temporal,lea2017temporal}, Temporal Convolutional Network (TCN) learned a hierarchy of intermediate feature representations and formed an encoder-decoder framework. Combining a deep ResNet \cite{he2016deep} with an RNN network, recurrent convolutional network was proposed to identify the surgical phase from videos \cite{jin2017sv,zisimopoulos2018deepphase}. These methods concentrate on frame-wise accuracy, however, segment-level performance is not fully focused which is limited by their training loss functions \cite{liu2018deep}. 

Recently, deep reinforcement learning (RL) was applied to gesture recognition. An agent learned its policy by interacting with the environment, i.e., surgical data, and achieved a state-of-the-art segment-level performance \cite{liu2018deep}. Since the agent relies on a neural network to produce decisions, the network confusion \cite{jin2017sv,Itzkovich2019using} problem remains unsolved for some rarely appeared gestures. Considering information from different stages might be a promising solution to this problem. This group of works target at generating more robust decisions than direct outputs for sequential decision problems. In \cite{silver2016mastering,silver2017mastering}, the capacity of AphaGo was enhanced by a large margin over the raw network, since Monte Carlo tree search (MCTS) \cite{coulom2006efficient} provides a look-ahead approach to bring statistics from future states. Although MCTS was designed for two-player games originally, Schadd \textit{et al}. raised single-player MCTS, where the average and best scores were both considered \cite{schadd2008single}.

In this paper, we come up with a novel and generic reinforcement learning framework for surgical video segmentation and classification through tree search. The proposed method is a search-based algorithm which produces current decisions by looking ahead into the future. We claim that it is only necessary to search for the frames that the policy agent produces uncertain decisions, which saves a lot of computing time. Hence, we design a gateway component to decide when to think carefully with the value network. More specifically, if the agent feels very confident about the output decisions, i.e., the maximum probability is above a threshold, decisions from a policy network are directly used to segment the input sequences; if not, tree search is invoked by considering the outputs from a policy and a value network. Our main contributions are summarized as follows: 
\begin{itemize}
	\item We propose a novel reinforcement learning based framework which performs in a human-like manner to generate decisions by jointly leveraging a policy network and a value network.
	\item We present a new tree search algorithm for decision refinement by potentially considering the future frames. This is crucial for accurate prediction of surgical gesture in online mode.
	\item We evaluate our proposed method on the public robotic surgery dataset of JIGSAWS. Our agent outperforms state-of-the-art results on surgical gesture recognition.
\end{itemize}

\begin{figure*}[t]
	\centering
	\includegraphics[width=0.8\textwidth]{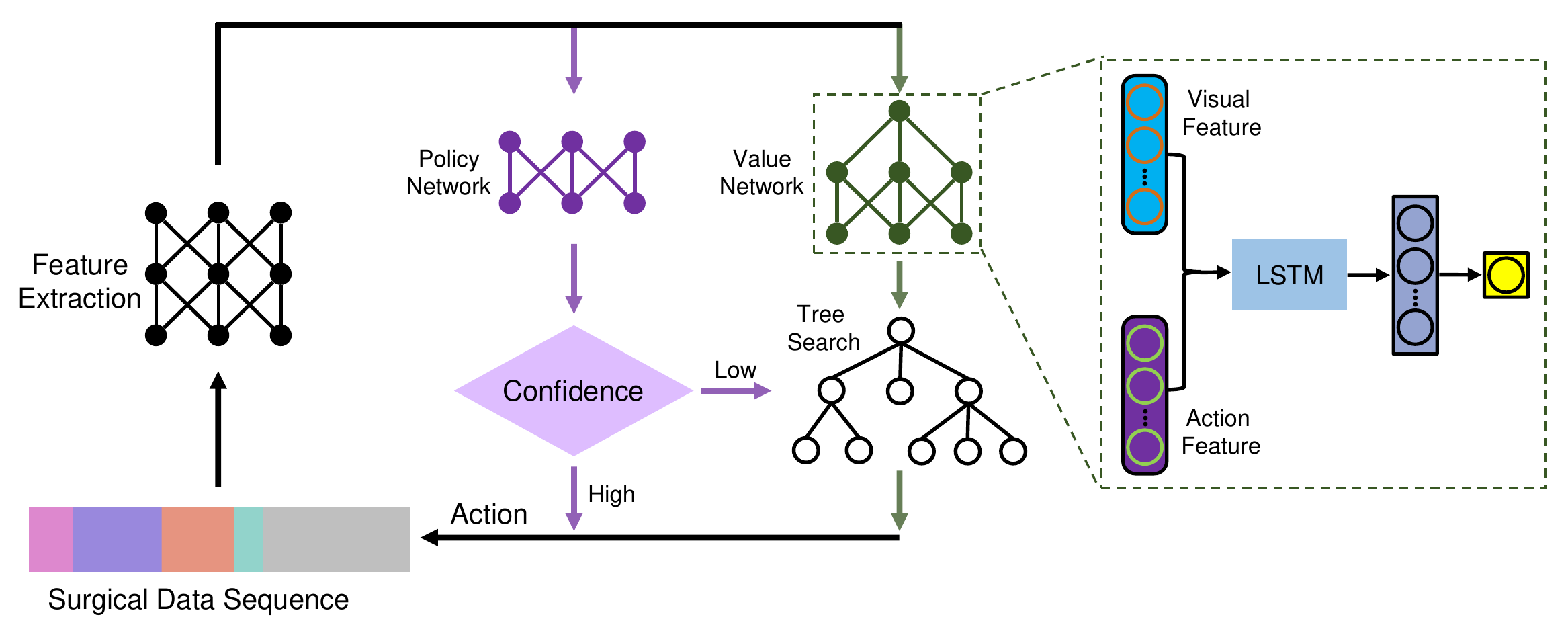}
	\vspace{-3mm}
	\caption{Overview of our proposed deep reinforcement learning based method for automatic surgical gesture recognition. Our framework consists of a policy network and a value network with tree search, which work together in a complementary manner.}
	\label{operate_pipeline}
	\vspace{-3mm}
\end{figure*}

%The rest of the paper is organized as follows. Section \ref{PRELIMINARY} provides some prerequisite knowledge. Section \ref{METHODS} presents the problem setup and introduces the process to conduct tree search. Section \ref{EXPERIMENTS} evaluates the performance of our framework on JIGSAWS dataset and gives some ablation analysis. Section \ref{DISCUSSIONS} discusses the potential applications of our method. Section \ref{CONCLUSIONS} draws the conclusion.
%======================================================
\section{PRELIMINARY}
\label{PRELIMINARY}
\subsubsection{Reinforcement learning} RL is about an agent interacting with the environment and learning to behave tactfully according to rewards in essence. This sequential decision problem can be formalized into a finite Markov Decision Process (MDP) \cite{sutton2018reinforcement} where the sets of states, actions, and rewards have a finite number of elements. At each step $t$, the agent faces some state $S_t\in \mathcal{S}$ and selects an action $A_t\in \mathcal{A}(s)$ based on $S_t$. One time step later, the agent receives a numerical reward $R_{t+1}\in \mathcal{R}\subset \mathbb{R}$. Then, the environment transfers to a new state $S_{t+1}$ at probability $p(S_{t+1}|S_t,A_t)$ and the agent wants to learn the optimal policy $\bm{\pi}(S_t)$ which is a distribution towards action set. The trajectory of agent is $S_0,A_0,R_1,S_1,A_1,R_2,...$ The goal is to maximize the accumulated rewards in an episode.

\subsubsection{Monte Carlo tree search} MCTS grows a search tree by asserting newly gained information asymmetrically \cite{browne2012survey}. It includes four steps in each simulation process \cite{coulom2006efficient}: selection, expansion, simulation, and backup. Firstly, the most urgent node is selected by a tree policy. Then, one or more child nodes are expanded by the selected node. Thirdly, a complete episode is run under rollout policy from one of its newly-added child nodes. Eventually, the simulation outcome is used to update the statistics of its ancestors. After a certain number of simulations, the next action can be chosen according to the statistics from the simulated outcomes.	

%======================================================	
\section{METHODS}
\label{METHODS}

In this section, we formulate our problem using an MDP model \cite{liu2018deep}, in which the agent regards the visual features as environment states and 
%segmentation operations 
surgical gesture recognition as decisions.
The Fig.~\ref{operate_pipeline} depicts an overview of our framework which is mainly consisted of a policy network and a value network. These two networks work together in a hierarchical way. We will describe each component in details, and particularly for the proposed tree search algorithm which is key to our framework.
%Following the description in Fig. \ref{operate_pipeline}, we introduce each component of the proposed method and the search algorithm in detail.

\subsection{Problem setup}

To segment and recognize the surgical gestures from the video data, the agent starts at the beginning of the video sequence $\{x^t\}_{t=0}^{T}$ and moves towards the end of the sequence. Based on the observation at some position, i.e., the environment state, the agent selects an action $a_i=(k,c)$ consisting of a step size $k$ and a gesture class $c$ for the frames stepped over. The length of $k$ can choose a small step $k_s$ or a large step $k_l$ based on the confidence in the gesture prediction to give. When the agent goes over all the data sequence $\{x^t\}_{t=0}^{T}$, the episode ends and this trajectory can be evaluated by user-preferred criteria.

This problem can also be regarded as a path-finding problem that the agent wants to give a sequence of actions to get the maximum scores given pre-defined criteria. At each state, there are $b$ actions to be chosen from and the complexity is exponential. In fact, with the ground truth label sequence $\{y^t\}_{t=0}^{T}$, the optimal path is definite. In this paper, the proposed approach helps the agent consider the best path to go ahead.

A high-quality feature base is crucial for taking good advantage of our method.
In this regard, we exploit the high-level representative TCN features as the input of our neural networks~\cite{lea2016temporal}. Specifically, the video data are first processed by a spatial CNN \cite{lea2016segmental} to generate raw features. 
Then, we re-train the TCN model with  modifying the original loss to the weighted cross-entropy loss~\cite{liu2018deep}. 
Finally, the TCN features $\{s_{tcn}^t\}$ can be obtained from the last hidden layer of our well-trained TCN model.
	
\subsection{Policy network}

The policy network takes input as the environment state and outputs a distribution over action space \cite{liu2018deep}. Then, the observation at frame $t$ is defined as
\begin{eqnarray}
s_p^t := (s_{tcn}^t,s_{tcn}^{t+k_s},s_{tcn}^{t+k_l},s_{trans},s_{hot}),
\end{eqnarray}
where $s_{trans}$ are probabilities from a statistical language model \cite{richard2016temporal} and $s_{hot}$ is a one-hot vector indicating the gesture class given by the last action. The reward for each action is given by
\begin{eqnarray}
r(s^t_p,(k,c)) :=\alpha k-\sum_{t'=t}^{t+k-1}\textbf{1}(y^{t'}\neq c) ,
\end{eqnarray}
where $\alpha$ is a weight parameter. This reward definition incites larger steps while penalizes wrong predictions. Then, the policy is optimized by using the Trust Region Policy Optimization \cite{schulman2015trust}. 

\subsection{Value network}
The value network takes as an input the representation of the environment state and the action choice of the agent, and outputs an advantage score of each action. The observation at frame $t$ is defined as
\begin{eqnarray}
s_v^t := (s_{tcn}^t,s_{hot}^t),
\end{eqnarray}
where $s_{hot}^t$ is a one-hot vector indicating the conjectured gesture class of the frame $t$. Note that $s_{hot}^t$ and $s_{hot}$ share the same kind of representation. Here, we regard $\{s_v^t\}$ as a new sequence data to jointly consider the surgical data and candidate gesture categories, which is then sent to a recurrent neural network. The reward for each action is 
\begin{eqnarray}
r(s^t_v,(k,c)) :=\sum_{t'=t}^{t+k-1}{\textbf{1}}(y_{t'}= c)-
\sum_{t'=t}^{t+k-1}{\textbf{1}}(y_{t'}\neq c) .
\end{eqnarray}
Thus, the range of the global mean reward is $[-1,1]$.

The sketchy structure of the value network is demonstrated in Fig. \ref{operate_pipeline}. The input layer of the value network receives a sequence vector by concatenating TCN feature and its possible class. It connects to one hidden layer of LSTM and a fully connected layer, both with 32 neurons. In the output layer, we use a Tanh nonlinearity to produce a scalar.

Since the ground truth label sequence is available, we can directly employ supervised learning approach to train the network. The expert experiences are created with the ground truth labels. To be mentioned, we utilize training data augmentation to alleviate overfitting, in which the agent generates non-expert predictions using random strategy. 

Until an episode ends, the overall mean reward $\bar r^*$ for each frame $t$ is obtained and the data is stored as $(s^t_v,\bar r^*)$. Randomly chosen sequences with length $K$ are sent to LSTM \cite{jin2017sv} and the mean square loss function is optimized by Adam optimizer \cite{kingma2014adam}. To establish longer dependence over each frame, the model is trained with increasing $K$ \cite{michalski2014modeling}.

\begin{figure*}[t]
	\centering
	%\vspace{-1mm}
	\includegraphics[width=0.85\textwidth]{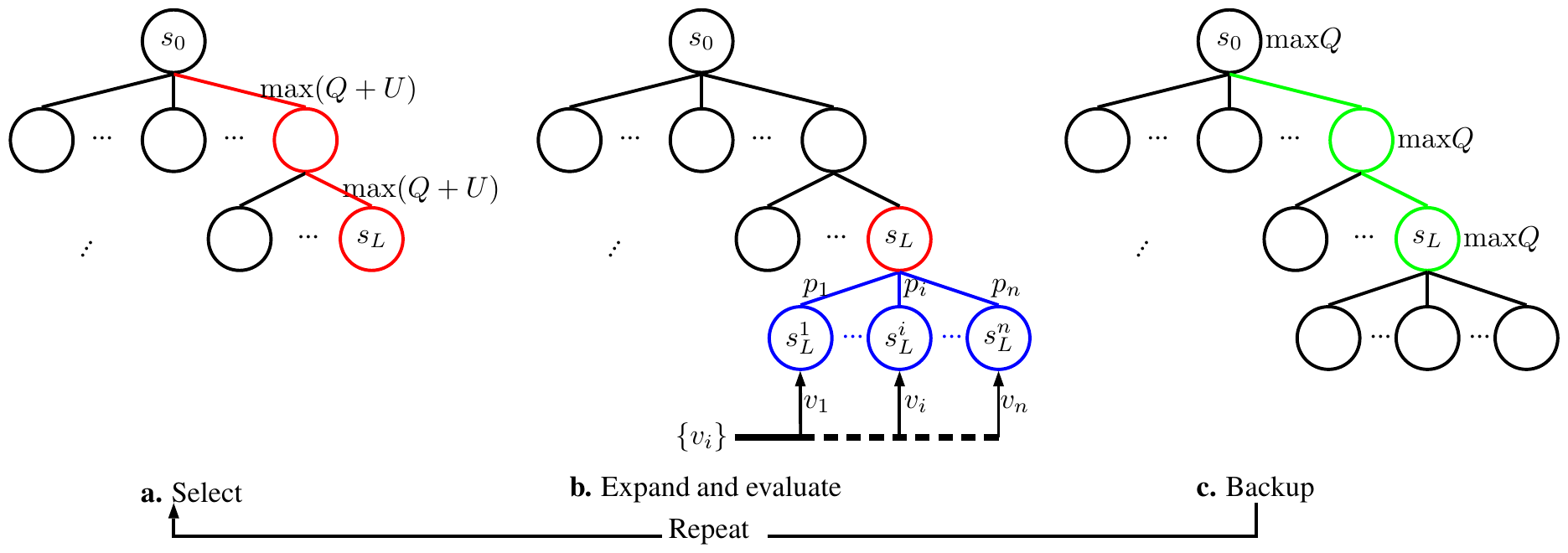}
	\caption{Tree search pipeline. \textbf a. Each search traverses the tree following the selection criterion from the root node. \textbf b. The leaf node $s_L$ expands all its child nodes and the prior probabilities are delivered to child nodes; each child node is evaluated to store its state value. \textbf c. All action values of the nodes in this search trajectory is updated to reflect the maximum value in its branches.}
	\label{aTreeSearch}
	\vspace{-3mm}
\end{figure*}

\subsection{Tree search algorithm}
We design a single-player tree search algorithm to fuse the outputs from a policy network and a value network. Since the environment behavior is deterministic, the constructed search tree does not contain the nodes for the environment. The purpose of the tree search is to return the best possible path starting from the current frame and help the agent make decisions.  
Since the prior probability of each child node is conditional on its parent node, a tree structure is applied to preserve this information. To facilitate the retrieval of the best path in the tree, we design a tree structure that each parent node preserves the maximum action value among all its child nodes. It is trivial that the node with maximum value can be chosen greedily from the root node.

The prior probabilities of child nodes are computed using a policy network:
\begin{eqnarray}
\bm{p}(s)=f_p(s_p^t).
\label{eqn:policy}
\end{eqnarray}
Due to the step length $k$ of each action, we use the average of the $k$ sequential outputs from the value network $f_v$ as the estimation of the global mean reward, i.e.,
\begin{eqnarray}
\bar r(s) = \frac{1}{k}\sum f_v([s_v^{t}:s_v^{t+k}]),
\label{fv}
\end{eqnarray} 
where $[s_v^{t}\hspace{-1mm}:\hspace{-1mm}s_v^{t+k}]$ is the observation sequence from current frame to the next $k$th frame.
If $s$ is a leaf node, its state value is calculated by 
\begin{eqnarray}
v(s) = \mathrm{mean}\{\bar r(s')\}, ~\mathrm{for~} s'\mathrm{~in~the~path~}s_0\to s,
\label{fvv}
\end{eqnarray}
where $s_0$ is the root node. Each leaf node re-evaluates the global mean reward by averaging all the exiting estimations from the same path because average lowers the variance. Because of the tree property stated above, the action value of each node is equivalent to
\begin{eqnarray}
Q(s,a) = \max_{s'|s,a\to s'} v(s'),
\end{eqnarray}
where $s,a \hspace{-1mm} \rightarrow \hspace{-1mm} s'$ indicates that a path eventually reached a leaf node $s'$ after taking an action $a$ from $s$. Actually, each node stores the estimation of the optimal path through itself.

The overall tree search process is illustrated as follows. In Fig. \ref{aTreeSearch}-a, the selection starts from the root node $s_0$ and ends until a leaf node $s_L$ is encountered. At each of these steps, an action with the maximum sum of $Q(s,a)$ and upper confidence bound (UCB) is selected: %\cite{silver2017mastering}

\begin{eqnarray}
a' = \mathop{\arg\max}_{a} \left(Q(s,a)+ U(s,a)\right).
\label{eqn:puctselect}
\end{eqnarray}
The policy distribution from the policy network is added into $U(s,a)$ to help narrow the search space,
\begin{eqnarray}
U(s,a) = c_{\mathrm{puct}}\left( 1+p(s,a) \right) \frac{\sqrt{\sum_b N(s,b)}}{1+N(s,a)},
\label{eqn:Ucb}
\end{eqnarray}
where $c_{\mathrm{puct}}$ is a constant determining the level of exploration and $N(s,a)$ is the visit count of edge $(s,a)$. In this phase, the optimal information is used to guide the search due to the reachability of each state and UCB enforces the agent to consider the rarely visited nodes.

In Fig. \ref{aTreeSearch}-b, all child nodes $\{s^i_L\}$ of $s_L$ is expanded in the tree. The policy network evaluates $s_L$ to deliver prior probability $p_i$ to each child node $s_L^i$ by Eq. (\ref{eqn:policy}), while state values of the newly expanded nodes are generated using Eq. (\ref{fv}) and Eq. (\ref{fvv}). Note that the value network is utilized in a batch style and actions with the same step length are included in the same batch. After this phase, $s_L^i$ stores the evaluation $v_i$ of itself.
In Fig. \ref{aTreeSearch}-c, the action values of $s_L$ is given by:
\begin{eqnarray}
Q(s_L,a) = \max v(s_L^i),
\end{eqnarray}
and its ancestor nodes with each step $j<=L$ are updated in a backward pass by:
\begin{eqnarray}
Q(s_j,a_j) = \max_{s_{j+1}|s_j,a_j\to s_{j+1}} Q(s_{j+1},a_{j+1}),
\label{eqn:backup}
\end{eqnarray}
and the visit count $N(s,a)$ of the each edge in this path is also increased. 
	
When a certain number of simulations are implemented, the action in the first edge with most visit times from $s_0$ is chosen. If more than one action have the same maximum visit count, the edge with the maximum action value is chosen. 

In our search algorithm, we do not choose a big simulation times to make $U(s,a)$ approach zero because the dependence of frames decays with the increasing distance. Thus, we let $p(s,a)$ always play a role in the selection phase and help prune away some inferior branches. Furthermore, our method ensures that each leaf node is judged by prior probability and value together. The pseudo code of our method describes the overall tree search process, as shown in Algorithm~\ref{algorithm1}.

%======================================================
\section{EXPERIMENTS}
\label{EXPERIMENTS}
In this section, we evaluate our proposed deep reinforcement learning method for surgical gesture recognition on the popular public JIGSAWS~\cite{gao2014jhu,ahmidi2017dataset} dataset.
We design experiments to answer the following two questions: 1) Does the tree search algorithm produce a better testing outcome? 2)  What role does each component in the framework play in the performance boost?
\subsection{Dataset}
We use the JIGSAWS, a public dataset captured by the \textit{da Vinci} Surgical System (dVSS, Intuitive Surgical Inc., CA, USA). It consists of video and kinematic data
from eight surgeons in three different levels of robotic surgical experience. The manual annotations describing the ground truth gesture classes for each frame are available. We use the video data from the suturing task with total 10 different gestures, i.e., reaching for the needle with right hand (G1), positioning the tip of the needle (G2), etc. There are 39 sequences in total and lengths are a few minutes.

\begin{algorithm}[t]
	\caption{Tree search algorithm} \label{algorithm1} 
	\begin{algorithmic}[0]  
		\Require Current states, search times $n$
		\Ensure Refined action $a^*$
		
		\State Initialize the root node $s_0$ according to current states	
		\For{$i = 1 , n$} 
		\State Start from $s_0$
		\State Go through the tree using Eq. (\ref{eqn:puctselect}) until a leaf node $s_L$ 
		\If {$s_L$ is not end state}
		\State Expand all its child nodes 
		\State Evaluate $s_L$ to output $\{p_i\}$ using Eq. (\ref{eqn:policy}) 
		\State Compute $\{v_i\}$ of its child nodes using Eq. (\ref{fv}--\ref{fvv})		
		\EndIf
		\State Update visit count of these nodes
		\State Update action values of the visited nodes by Eq. (\ref{eqn:backup}) 
		
		\EndFor 
		\If {only one $N(s_0,a)$ is maximum}
		\State $a^* \gets \arg \max\limits_{a} N(s_0,a)$  
		\Else
		\State $a^* \gets \arg \max\limits_{a} (N(s_0,a)+Q(s_0,a))$ 
		\EndIf
		\State \Return{$a^*$}

	\end{algorithmic}  
	\label{algorithm}
\end{algorithm}

\subsection{Evaluation metrics}
We examine three different evaluation metrics for different approaches: (i) Accuracy, i.e., the percentage of correctly recognized frames in a video. (ii) Edit score \cite{lea2016segmental}, the normalized Levenshtein distance between predicted gesture sequence and ground truth. (iii) F1@k score \cite{lea2017temporal} with different thresholds. It penalizes over-segmentation errors while does not for minor temporal shifts between the predictions and ground truth. Under this criterion, each predicted gesture segment is considered true positive if its Intersection over Union (IoU) towards the corresponding ground truth is above the threshold and vice versa. Then, F1 score is computed using precision and recall by: $\text{F1}=2\frac{prec*recall}{prec+recall}$.

\subsection{Implementation details}

We follow the leave-one-user-out (LOUO) setup for cross-validation, which is the same as used in \cite{ahmidi2017dataset}. And one model is trained for each experiment with one left-out user. The final evaluation metrics are calculated for each video in the test dataset and then averaged. A TCN is trained to generate features for the policy and value network. 

As a baseline for the proposed approach, we re-implement the method in \cite{liu2018deep} as policy network with one minor change, where actions are chosen deterministically, i.e., the action with maximum probability is executed, rather than using a stochastic strategy in testing stages. As for the value network, we also adopt the same parameters for $(k_s,k_l)$ and $\alpha$ as the policy network, i.e., (4, 21) and 0.1. As there are 10 gesture classes in this dataset, the search tree expands 20 child nodes each time. The length $K$ of training sequences increases from 20 to 100 with an interval of 10. Half of the training data for the value network are generated using random actions. We set the threshold of conducting tree search as 0.98. The constant $c_{\mathrm{puct}}$ and search times are set to 1.5 and 10 respectively.

\subsection{Experimental results and ablation analysis}

We perform experiments to test the abilities of the policy network, value network, and our combined network. The action with the biggest probability output by the policy network is directly used to segment an episode. Note that the tree search algorithm cannot be carried out with only the policy network because its output is conditional probability precisely. By contrast, the value network gives the estimated rewards for all the elements in the action set and the action with the maximum estimation is executed. Since the relationships between the current frame and future frames are limited and indefinite, we apply different search times for the proposed method. In fact, with only the value network, tree search can also be realized by assigning each action an equal prior probability. Thus, we also evaluate the tree search model with information only from the value network. 

\subsubsection{Testing of each component}
Table \ref{tab:search_times} summarizes the segmentation results respect to different search times. When searching 10 times for each consideration, the combined method achieves the highest segment-level edit score at a negligible cost of accuracy. Combining the policy and value network always outperforms the two individual networks concerning the frame-level accuracy. Also, the results of the value network indicate the importance of the policy network. The scores are not always going up with the increment of search times, which implies the limited effectiveness of the temporal information. Although it seems that the decisions of past frames are independent of the future choice of action, results of the value network show that they do help to choose a right class, at least on a segment level. By the way, the value network trained by supervised learning also achieves reasonable performance since this is a special reinforcement learning problem with expert data available.

\begin{table}[t]
	%\setlength{\abovecaptionskip}{0.1cm}
	%\setlength{\belowcaptionskip}{-0.2cm}
	%\small
\caption{Ablation experiments on tree search \protect \\ with different search times.}
	\label{tab:search_times}
	\centering
	\begin{tabular}{c|cccccc}
		\hline
		\multirow{2}{*}{Search times} & \multicolumn{2}{c}{Policy} & \multicolumn{2}{c}{Value} & \multicolumn{2}{c}{Policy+Value} \\
		& Acc          & Edit        & Acc         & Edit        & Acc             & Edit           \\ \hline
		0                             & 81.52        & 87.87       & 80.91       & 88.34       & 81.67           & 87.94          \\
		10                            & \multicolumn{2}{c}{-}      & 80.99       & 87.74       & 81.67           & 88.53          \\
		20                            & \multicolumn{2}{c}{-}      & 80.99       & 87.74       & 81.70           & 88.24          \\
		30                            & \multicolumn{2}{c}{-}      & 81.02       & 88.16       & 81.61           & 88.22          \\
		40                            & \multicolumn{2}{c}{-}      & 81.01       & 88.16       & 81.69           & 87.86          \\ \hline
	\end{tabular}
	\vspace{-3mm}
\end{table}

\subsubsection{Behaviors of the policy network}
\begin{figure}[t]
	\centering
	\includegraphics[width=0.45\textwidth]{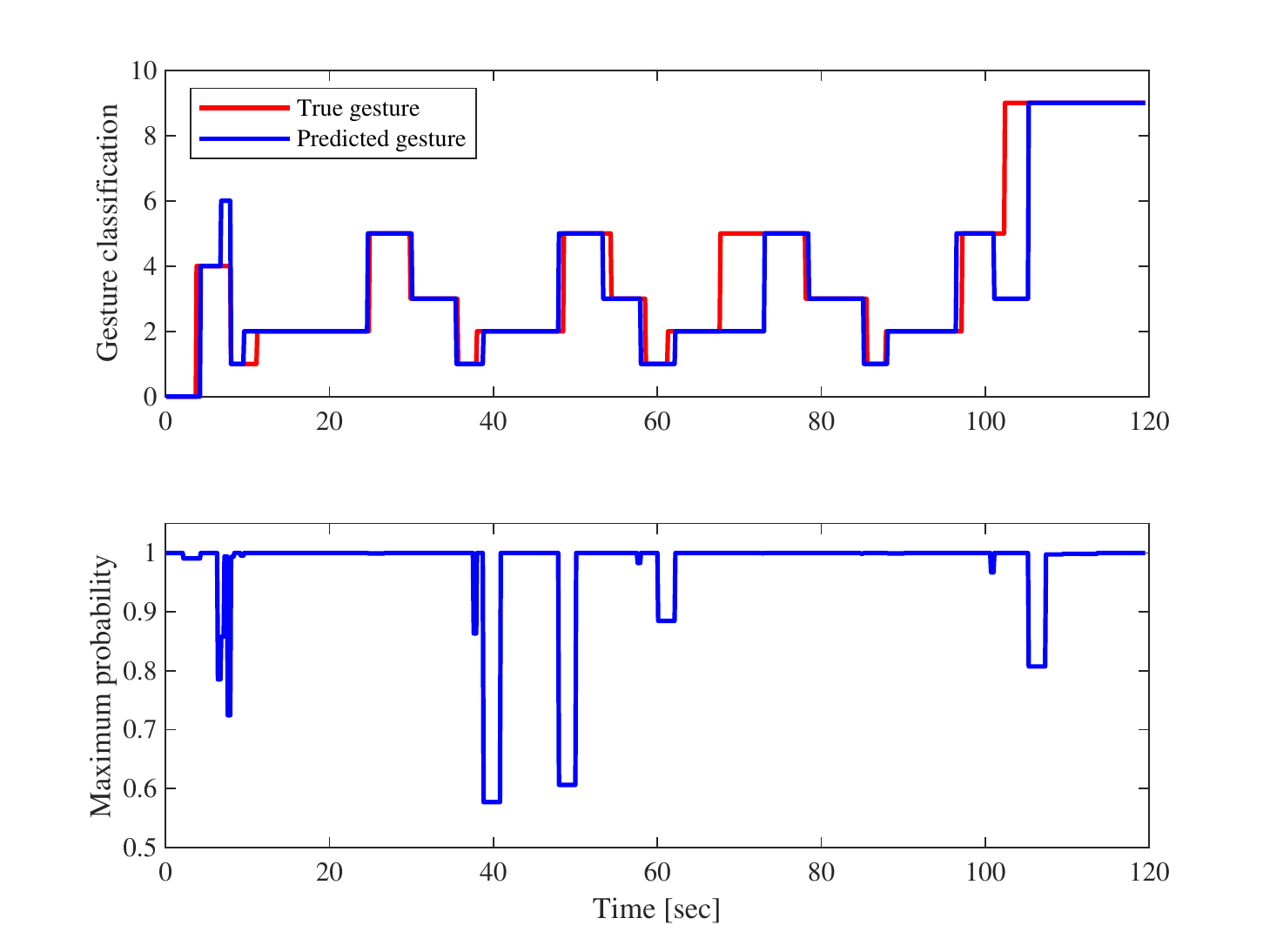}
	\vspace{-4mm}
	\caption{In the upper part, we present the recognition results from the policy network and ground truth in one complete video; in the lower part, the corresponding predicted probabilities are shown which are used to choose the executed action.}
	\label{prediction_example}
	\vspace{-3mm}
\end{figure}

\begin{figure}[b]
	\vspace{-2mm}
	\centering
	\includegraphics[width=0.45\textwidth]{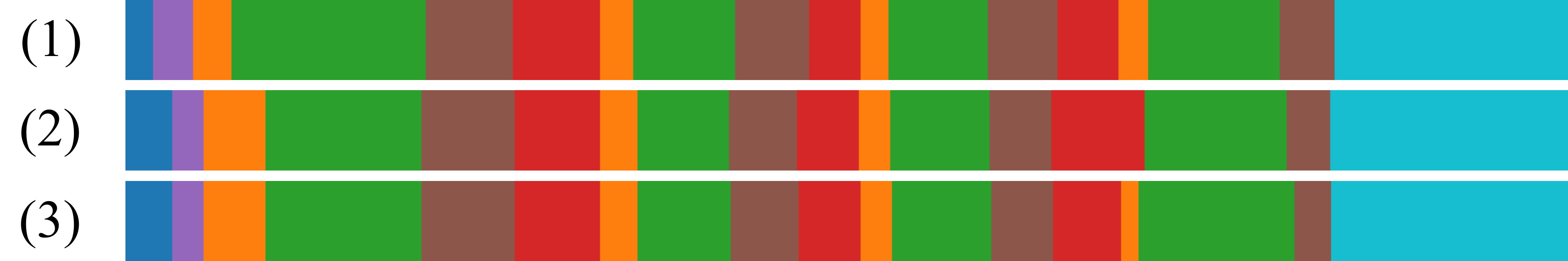}
	%\vspace{-1mm}
	\caption{Color-coded ribbon illustration of surgical gesture from a complete video. %whose horizontal axis represents the time progression. 
	We present (1) the ground truth (2) recognition results from the policy network (3) rectified predictions by tree search. Each color stands for a gesture class. The method based on tree search could recognize the missing gesture of the policy network.}
	\label{bar_example}
\end{figure}

To inspect the detailed behavior of the policy network, a prediction example generated is visualized. As shown in Fig. \ref{prediction_example}, the predicted classes follow the trend of ground truth (Please refer to \cite{ahmidi2017dataset} for gesture classification numbers). We also plot the maximum probability for each frame to examine its confidence in an episode. It is interesting that the policy network tends to be ambiguous at gesture boundaries. Most of the time, it is certain to make the right decision although it misses the right actions for some parts with high confidence. Since the tree search is only applied at the uncertain frames, the improvement relative to the policy network mainly depends on the more accurate recognition of boundaries. The over-segmentation problem is also alleviated as the edit score is raised. Furthermore, we show a prediction example of the segmentation outcomes of direct decisions by the policy network and rectified decisions using tree search. As shown in Fig. \ref{bar_example}, one gesture is missed in the second half of the episode by the policy network. By contrast, the missing gesture is recognized by our proposed method.

\subsubsection{Overall performances}
Table \ref{tab:outcome} shows the experimental results on video data. Our tree search based method is compared with the original TCN, RL based method, and other recent works. The evaluation metrics are accuracy, edit score and F1 score with the IoU threshold set to 10\%, 25\%, 50\% respectively. Compared with existing works, our approach achieves higher scores given all evaluation criterion. The reproduced outcome using RL based method attains almost the same performances as reported in \cite{liu2018deep}. For the value network, its performance is obtained without tree search. Interestingly, the policy network enjoys a superior ability on accuracy than the value network, while the value network behaves better on edit score. Through tree search for 10 times, the two cooperate to create an even better outcome. The processing time for the 10 times search setting and \cite{liu2018deep} are 25.4 s and 6.2 s, respectively, which are completely enough for an online mode given the total length of videos (73 mins). The additional time of our method is caused by re-considerations of 13\%  frames.

% JIGSAWS

\section{DISCUSSIONS}
\label{DISCUSSIONS}
The proposed method is a search-based algorithm which makes it more generic for decision problems. The problem formulated using RL framework here is a special case that the agent's actions do not affect the transfer of the environment. Therefore, the interactions of adjacent actions are quite limited. Our method makes the agent act in a human-like manner by jointly leveraging two networks. The policy network gives quick decisions and the value network offers meticulous selections. The intuition is that the value network could rectify the policy network's faults by providing advice from a different perspective. 

\begin{table}[t]
	%\setlength{\abovecaptionskip}{0.1cm}
	%\setlength{\belowcaptionskip}{-0.2cm}
	%\small
	\caption{Results on the suturing task of JIGSAWS.}
	\label{tab:outcome}
	\centering
	\begin{tabular}{c|ccccc}
		\hline
		Method                              & Acc            & Edit           & \multicolumn{3}{c}{F1@\{10,25,50\}}              \\ \hline
		MsM-CRF \cite{tao2013surgical}     & 71.78          & -              & \multicolumn{3}{c}{-}                            \\
		Seg-ST-CNN \cite{lea2016segmental} & 74.22          & 66.56          & \multicolumn{3}{c}{-}                            \\
		TCN \cite{lea2016temporal}         & 81.4           & 83.1           & \multicolumn{3}{c}{-}                            \\
		TCN+Deep RL \cite{liu2018deep}     & 81.43          & 87.96          & 92.0           & 90.5           & 82.2           \\ \hline
		Policy Net                          & 81.52          & 87.87          & 92.20          & 90.86          & 82.77          \\
		Value Net                           & 80.91          & 88.34          & 92.32          & 90.10          & 81.36          \\
		Policy+Value (\bf{Ours})                      & \textbf{81.67} & \textbf{88.53} & \textbf{92.68} & \textbf{90.99} & \textbf{83.15} \\ \hline
	\end{tabular}
	\vspace{-3mm}
\end{table}

Our work is novel in terms of formulating the surgical video gesture recognition task into a path searching problem. Rather than purely relying on a policy network as Liu and Jiang \cite{liu2018deep} recently investigated, we further introduce a value network into the framework, by borrowing the spirit of AlphaGo \cite{silver2016mastering,silver2017mastering}. More importantly, we develop a tree search algorithm associated with the value network for leveraging global information. In the problem setting of online prediction for the surgical gesture, future frames are not available to the system. By taking advantage of our proposed tree search algorithm, predictions of future frames can be explosively considered for helping make decisions on the current time step. As a nearly online mode, another alternative solution of missing future frames is to output slightly delayed gesture predictions, which makes the proposed tree search feasible.
In practice, such a value network functions together with the policy network, when a frame receives a less confident decision by the policy network. In other words, the value network is an inseparable module to compensate defective predictions from the policy network. We think our introduced insight will inspire more future studies on using reinforcement learning for surgical video analysis. For our future work, we plan to employ disparate features (e.g., visual and kinematic data) to train the two networks respectively, for further stimulating complementarity of the policy and value networks.

%======================================================	
\section{CONCLUSIONS}
\label{CONCLUSIONS}	
In this paper, we propose a novel method based on tree search for surgical video segmentation problems. The tree search algorithm unites the outputs from different neural networks. During tree search, prior probabilities from the policy network narrow down the search space and guide the search directions together with evaluations from the value network. Due to a more comprehensive consideration of action selections, the suggested approach outperforms the baseline methods as well as the existing works on JIGSAWS dataset in terms of different metrics. To conclude, we highlight the benefit of introducing the contemplation ability for the agent when getting confused about the direct decisions, which is of great importance in the medical field.
	
%\addtolength{\textheight}{-12cm}   % This command serves to balance the column lengths
	% on the last page of the document manually. It shortens
	% the textheight of the last page by a suitable amount.
	% This command does not take effect until the next page
	% so it should come on the page before the last. Make
	% sure that you do not shorten the textheight too much.
		
%\section*{ACKNOWLEDGMENT}

%The authors would like to thank Colin Lea for sharing raw features of JIGSAWS dataset. 
	
\newpage
	
\bibliographystyle{IEEEtran}
\bibliography{IEEEabrv}

\end{document}